%% file: main.tex
\documentclass{acmart}

\usepackage{times}
\usepackage{latexsym}

\usepackage[utf8]{inputenc}

\usepackage{microtype}

\usepackage{lineno} 
\usepackage{graphicx}



\usepackage{tikz}
\usetikzlibrary{shapes.geometric, arrows.meta, positioning, fit}
\usepackage{amsmath}
\usepackage{adjustbox}
\usepackage{booktabs}
\usepackage{stfloats}
\usepackage{subcaption}

\title{Categorical Prior Lock-in: Why In-Context Learning Fails for Structured Data}

\begin{document}

\author{Antonio Pelusi}
\orcid{0009-0006-0807-4364}
\affiliation{\institution{University of Insubria}\country{Italy}}

\email{apelusi3@uninsubria.it}
\author{Stefano Braghin}
\orcid{0000-0001-5519-1674}
\affiliation{\institution{IBM Research}\country{Ireland}}
\email{stefanob@ie.ibm.com}
\author{Alberto Trombetta}
\orcid{0000-0002-2567-9297}
\affiliation{\institution{University of Insubria}\country{Italy}}
\email{alberto.trombetta@uninsubria.it}

\input{abstract}

\maketitle

\input{introduction}
\input{related}
\input{experimental}
\input{evaluation}
\input{results}
\input{discussion}
\input{conclusion}

% \newpage
% \input{limitations}

\bibliographystyle{ACM-Reference-Format}
\bibliography{references}

\end{document}

%% file: abstract.tex
\begin{abstract}
Large language models (LLMs) are increasingly used as conditional generators for structured data, relying on in-context learning (ICL) to adapt to new distributions without parameter updates. We investigate the limits of ICL for structured generation under distribution mismatch, using high-cardinality tabular data as a controlled test case, and identify a structural failure mode we term \textit{categorical prior lock-in}: the inability of ICL to update the model's prior over token distributions inherited from pre-training. Across two 7B-parameter open-weight models, ICL improves numerical fidelity with additional examples but exhibits a sharp ceiling on categorical distributions, failing to reproduce rare classes entirely. Parameter-efficient fine-tuning (LoRA) overcomes these limitations but introduces measurable memorization risk and, in some cases, destabilizes structured output generation, highlighting a fundamental trade-off between adaptability and privacy.
\end{abstract}

%% file: introduction.tex
\section{Introduction}
\label{sec:intro}
Large language models (LLMs) have demonstrated a remarkable ability to perform new tasks from a small number of examples provided in context, without updating model parameters. This paradigm, known as in-context learning (ICL), is often interpreted as a form of implicit adaptation, enabling models to approximate new input-output mappings through conditioning alone. As a result, ICL has been widely adopted as a lightweight alternative to fine-tuning, particularly in settings where parameter updates are costly or infeasible.
 
A growing line of work applies this paradigm to structured generation tasks, including the synthesis of tabular data, where the model is prompted with a schema and example records and expected to generate samples matching the target distribution. While recent studies report promising results, these evaluations primarily focus on downstream utility metrics and often overlook the underlying distributional behaviour of the model, leaving open a fundamental question: \textit{to what extent can ICL adapt a pre-trained language model to a new data distribution in structured generation tasks?}
 
In this work, we answer that question empirically and identify a structural failure mode we term \emph{categorical prior lock-in}: the tendency of a language model to adhere to its pre-training distribution over tokens, even when conditioned on examples drawn from a different target distribution. Using high-cardinality categorical features as a stress test, we show that ICL cannot approximate rare or domain-specific label distributions regardless of prompt design or the number of in-context examples. Unlike numerical features -- whose fidelity improves consistently as more examples are provided -- categorical distributions exhibit a sharp ceiling, and minority categories are never reproduced. This limitation propagates to inter-feature dependencies and minority class omission, and is not resolved by prompt engineering within the tested range. We analyse this phenomenon and its mechanisms in depth in Section%~\ref{sec:prior-lock-in}.
~\ref{sec:discussion}.
 
In contrast, parameter-efficient fine-tuning (LoRA) substantially improves both marginal and joint fidelity by performing the global distributional adjustment that ICL cannot. This improvement, however, comes at a cost: all fine-tuned configurations exhibit proximity to training records below the memorization threshold, raising concerns about data leakage and privacy. Our contributions are as follows: (1) we identify and characterise categorical prior lock-in as a fundamental failure mode of ICL in structured generation; (2) we provide a systematic empirical evaluation across models and prompting strategies; and (3) we analyse the trade-offs between ICL and parameter-efficient fine-tuning in terms of fidelity, structural consistency, and privacy risk.

%% file: related.tex
\section{Related Work}
\label{sec:related}

The dominant pre-LLM approach relies on Generative Adversarial Networks. CTGAN~\cite{xu2019modelingtabulardatausing} addresses mixed column types and class imbalance through a conditional generator and mode-specific normalization, establishing a widely-used baseline. GAN-based methods, however, require explicit architectural decisions per column type and struggle with high-cardinality categoricals, a limitation LLM-based methods were expected to overcome through pre-trained semantic knowledge.

Brown et al.~\cite{brown2020languagemodelsfewshotlearners} established that LLMs can perform novel tasks from a handful of demonstrations without weight updates, a capacity extended to tabular synthesis by subsequent work. CLLM~\cite{seedat2024curatedllmsynergyllms} showed that LLMs prompted with in-context examples, paired with a post-hoc curation mechanism, can augment tabular datasets in low-data regimes without fine-tuning. EPIC~\cite{kim2025epiceffectivepromptingimbalancedclass} systematically characterized prompt design elements for ICL-based generation, showing that balanced, grouped examples mitigate class imbalance. Both works, however, focus on frontier models and report ML utility metrics only, leaving open questions about distributional fidelity on high-cardinality categoricals and privacy risk under local deployment.

GReaT~\cite{borisov2023languagemodelsrealistictabular} first demonstrated that fine-tuning a pre-trained autoregressive model on serialized tabular records produces realistic synthetic data. REaLTabFormer~\cite{solatorio2023realtabformergeneratingrealisticrelational} extended this to relational datasets, and TabuLa~\cite{zhao2025tabulaharnessinglanguagemodels} explored efficiency improvements for the same paradigm. None of these works, however, targets organizations under data residency constraints, for whom 7B instruction-tuned models with parameter-efficient adaptation represent the upper limit of on-premise deployment.

%% file: experimental.tex
\section{Experimental Setup}
\label{sec:experimental}

% We here present in details the empirical assessment of our approach

\paragraph{Dataset}

The dataset is a publicly available credit card transaction log with binary fraud labels. Each record contains seven numerical features,transaction amount (\texttt{amt}), cardholder and merchant geographic coordinates (\texttt{lat}, \texttt{long}, \texttt{merch\_lat}, \texttt{merch\_long}), city population (\texttt{city\_pop}), and Unix timestamp (\texttt{unix\_time}), and four categorical features: transaction category (\texttt{category}), gender (\texttt{gender}), occupation (\texttt{job}), and US state (\texttt{state}). The \texttt{job} field contains 494 distinct labels with a highly imbalanced distribution (most common label: 0.75\% of records). Geographic coordinates are constrained to continental US territory with strong inter-feature correlations ($r > 0.99$ between cardholder and merchant coordinates). \texttt{amt} and \texttt{city\_pop} follow heavy-tailed distributions (skewness $= 42.3$, kurtosis $= 4545.6$; median $= 2456$ vs.\ mean $= 88824$). Only 0.58\% of transactions are fraudulent, a ratio critical for downstream classifier calibration.

\paragraph{Models}

\textbf{Qwen2.5-7B-Instruct}~\cite{qwen2025qwen25technicalreport} and \textbf{Mistral-7B-Instruct-v0.3}~\cite{jiang2023mistral7b} are open-weight, instruction-tuned decoder-only transformers selected for local deployment under data residency constraints. Their differences in training data and methodology allow us to assess whether observed behaviors are model-specific or general to this scale.

\paragraph{Generation Strategies}

All configurations use a schema-encoding base prompt across three paradigms: \textbf{Zero-shot ICL}; \textbf{Few-shot ICL} (1, 5, or 10 records); and \textbf{LoRA fine-tuning}~\cite{hu2021loralowrankadaptationlarge} (trained on 10\% or 50\% of real data, then prompted zero-shot).

\paragraph{Generation Pipeline and Validation}

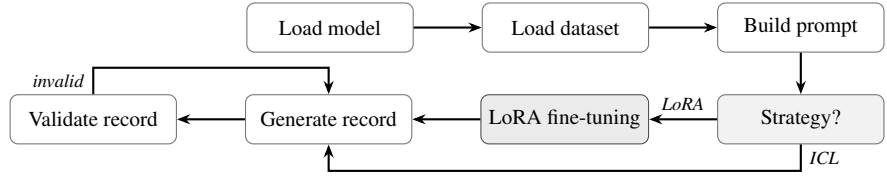
\begin{figure}[tb]
    \centering
    \begin{adjustbox}{max width=\columnwidth}
    \begin{tikzpicture}[
        node distance = 0.55cm and 0.9cm,
        box/.style      = {rectangle, rounded corners=3pt, draw=black!55,
            minimum width=2.2cm, minimum height=0.65cm,
            align=center, font=\small, fill=white},
        decision/.style = {box, fill=black!5},
        highlight/.style= {box, fill=black!8, draw=black!70},
        arr/.style      = {-{Stealth[length=5pt]}, thick},
        lbl/.style      = {font=\footnotesize},
        ]
        \node[box]      (model)   {Load model};
        \node[box,      right=of model]   (data)    {Load dataset};
        \node[box,      right=of data]    (prompt)  {Build prompt};
        \node[decision, below=of prompt]  (strat)   {Strategy?};
        \node[highlight,left=of strat]    (lora)    {LoRA fine-tuning};
        \node[box,      left=of lora]     (gen)     {Generate record};
        \node[box,      left=of gen]      (val)     {Validate record};
        \draw[arr] (model)  -- (data);
        \draw[arr] (data)   -- (prompt);
        \draw[arr] (prompt) -- (strat);
        \draw[arr] (strat)  -- node[lbl, above]{\textit{LoRA}} (lora);
        \draw[arr] (lora)   -- (gen);
        \draw[arr] (gen)    -- (val);
        \draw[arr]
        (strat.south)
        -- node[lbl, right]{\textit{ICL}}
        ++(0, -0.35)
        -| (gen.south);
        \draw[arr]
        (val.north)
        -- node[lbl, left]{\textit{invalid}}
        ++(0, 0.35)
        -| (gen.north);
    \end{tikzpicture}
    \end{adjustbox}
    \caption{Generation pipeline.}
    \label{fig:pipeline}
\end{figure}

As illustrated in Figure~\ref{fig:pipeline}, all configurations follow a unified pipeline\footnote{Code available at: \url{https://github.com/antoniopelusi/LLM-synthetic-data-generation/releases/tag/v1.0.0}}.

Records are generated one at a time via temperature sampling ($T = 0.8$, top-$p = 0.95$, repetition penalty $= 1.05$) with a fixed random seed. A record is valid if it parses as well-formed JSON, contains all required fields, and conforms to the expected type signature; invalid records are discarded and regenerated until 10000 valid records are collected.

%% file: evaluation.tex
\section{Evaluation Framework}
\label{sec:evaluation-framework}

\paragraph{Structural Reliability}

Structural reliability is measured by the error rate: the fraction of generation attempts that fail the validation pipeline, a key indicator of deployment efficiency as high failure rates impose significant computational overhead.

\subsection{Distributional Fidelity}

Statistical fidelity is assessed using Total Variation Distance (TVD) computed independently for each feature between the synthetic and real marginal distributions:
\begin{equation}
    \text{TVD}(P, Q) = \frac{1}{2} \sum_{x \in \mathcal{X}} |P(x) - Q(x)|
\end{equation}
TVD is bounded in $[0,1]$, where 0 indicates identical distributions and 1 completely disjoint support. Continuous features are discretized into 50 equal-width bins before computation. Note that TVD captures only marginal distributions: a synthetic dataset can match real marginals while failing to preserve inter-feature correlations, making TVD a necessary but not sufficient condition for distributional fidelity.

\paragraph{Privacy and Memorization Risk}

We quantify memorization risk via the DCR Ratio, which compares the average distance from each synthetic record to its nearest real neighbor against the average nearest-neighbor distance within the real dataset:
\begin{equation}
    \text{DCR Ratio} =
    \frac{\mathbb{E}_{s \sim S}\!\left[\min_{r \in R} d(s, r)\right]}
    {\mathbb{E}_{r \sim R}\!\left[\min_{r' \in R \setminus \{r\}} d(r, r')\right]}
\end{equation}
where $S$ and $R$ are the synthetic and real record sets, and $d(\cdot,\cdot)$ is the Euclidean distance over the six numerical features standardized via z-score normalization; categorical features and \texttt{unix\_time} are excluded. A ratio above 1.0 indica tes synthetic records are on average more distant from real records than real records are from each other; below 1.0 suggests the model may be reproducing training records too closely. As a mean-based statistic, the DCR Ratio can mask tail behavior; it should therefore be interpreted alongside the exact duplication rate---the fraction of synthetic records whose distance to the nearest real record falls below 0.01 in standardized feature space.

\paragraph{Fraud Class Reproduction}

Fraud class reproduction is assessed by comparing the prevalence of fraudulent transactions (\texttt{is\_fraud} $= 1$) in the synthetic corpus against the real base rate of 0.58\%. Complete absence of fraudulent records is treated as a categorical failure, as any classifier trained on such data will lack exposure to the minority class entirely.

\paragraph{Inter-Feature Correlation Preservation}

We assess inter-feature dependency preservation by computing the Pearson correlation matrix over the seven numerical features and the binary fraud label for both the real and synthetic datasets. Two summary statistics are reported: the maximum absolute pairwise correlation error and the total absolute correlation error over all off-diagonal entries, with lower values indicating better preservation of the dependency structure.

%% file: results.tex
\section{Results}
	
	\subsection{Structural Reliability}
	
	\begin{table}[ht]
		\centering
		\caption{Generation Reliability}
		\label{tab:reliability}
		\footnotesize
		\begin{adjustbox}{max width=\columnwidth}
			\begin{tabular}{llrrrr}
				\toprule
				\textbf{Model} & \textbf{Strategy} & \textbf{Valid} & \textbf{Errors} & \textbf{Error Rate (\%)} & \textbf{Total Attempts} \\
				\midrule
				Qwen2.5-7B & ICL Zeroshot       & 10,000 & 118   & 1.17\%  & 10,118 \\
				& ICL Fewshot 1r     & 10,000 & 3     & 0.03\%  & 10,003 \\
				& ICL Fewshot 5r     & 10,000 & 0     & 0.00\%  & 10,000 \\
				& ICL Fewshot 10r    & 10,000 & 1     & 0.01\%  & 10,001 \\
				& LoRA FT 10\%       & 10,000 & 0     & 0.00\%  & 10,000 \\
				& LoRA FT 50\%       & 10,000 & 0     & 0.00\%  & 10,000 \\
				\midrule
				Mistral-7B & ICL Zeroshot       & 10,000 & 372   & 3.59\%  & 10,372 \\
				& ICL Fewshot 1r     & 10,000 & 139   & 1.37\%  & 10,139 \\
				& ICL Fewshot 5r     & 10,000 & 90    & 0.89\%  & 10,090 \\
				& ICL Fewshot 10r    & 10,000 & 128   & 1.26\%  & 10,128 \\
				& LoRA FT 10\%       & 10,000 & 6,962 & 41.04\% & 16,962 \\
				& LoRA FT 50\%       & 0      & 1,000 & 100.00\%& 1,000  \\
				\bottomrule
			\end{tabular}
		\end{adjustbox}
	\end{table}
	
	Qwen2.5-7B behaves consistently across all tested configurations. The 1.17\% zero-shot ICL error rate---already low enough for most practical purposes---drops to near-zero with a single in-context example and reaches exactly zero at five. Fine-tuning introduces no degradation, with both LoRA configurations at 0.00\%.
	
	Mistral-7B's behavior under ICL is acceptable, if somewhat noisier: error rates fluctuate between 0.89\% and 3.59\% across the few-shot series. Fine-tuning tells a very different story. At 10\% data exposure, the error rate climbs to 41.04\%---over ten times the zero-shot ICL baseline. At 50\% exposure, generation fails completely, producing no valid records in 1000 attempts. The LoRA 50\% configuration is excluded from all subsequent analyses.
	
	The high rejection rates, especially for Mistral-7B, trace back to a range of parsing and structural anomalies we identified during validation. For Qwen2.5-7B, these rare failures are limited to minor syntactic inconsistencies in zero-shot settings, such as the injection of code snippets or arithmetic operators (e.g., the \texttt{+} symbol) within JSON fields, the representation of \texttt{cc\_num} as a string, and slight schema deviations like duplicating the \texttt{zip} field or mislabeling \texttt{Unnamed: 0} as \texttt{Unnamed}. 
	
	Mistral-7B exhibits more severe issues: infinite sequences in \texttt{trans\_num}, integers parsed from numerical strings, spurious quotation marks, and hallucinated parameters—all causing immediate validation rejection.
	
	In our runs, errors were noticeably less frequent when two conditions held:
	\begin{itemize}
		\item \textbf{Contextual Alignment:} The use of few-shot examples or LoRA fine-tuning provides the necessary grounding to align the models with the target schema.
		\item \textbf{Explicit Constraints:} The inclusion of explicit \texttt{constraints} within the prompt serves as a critical safeguard against formatting drift.
	\end{itemize}

	\subsection{Distributional Fidelity} \label{sec:distributional_fidelity}
	
	\paragraph{Numerical Distributional Fidelity} \label{sec:numerical_distributional_fidelity}
	
	\begin{table*}[tb]
		\centering
		\caption{Per-Feature Numerical TVD}
		\label{tab:numerical_tvd}
		\scriptsize
		\begin{adjustbox}{max width=\textwidth}
			\begin{tabular}{llrrrrrrr}
				\toprule
				\textbf{Model} & \textbf{Strategy} & \textbf{amt} & \textbf{city\_pop} & \textbf{lat} & \textbf{long} & \textbf{merch\_lat} & \textbf{merch\_long} & \textbf{unix\_time} \\
				\midrule
				Qwen2.5-7B & ICL Zeroshot    & 0.0506 & 0.4144 & 0.5183 & 0.6499 & 0.4565 & 0.6415 & 0.6931 \\
				& ICL Fewshot 1r  & 0.0021 & 0.1747 & 0.2642 & 0.3730 & 0.2733 & 0.4048 & 0.5884 \\
				& ICL Fewshot 5r  & 0.0127 & 0.0417 & 0.2179 & 0.2990 & 0.2175 & 0.3162 & 0.4829 \\
				& ICL Fewshot 10r & 0.1004 & 0.0453 & 0.2196 & 0.2937 & 0.1785 & 0.2937 & 0.3745 \\
				& LoRA FT 10\%    & 0.0073 & 0.0175 & 0.0432 & 0.0429 & 0.0403 & 0.0403 & 0.0767 \\
				& LoRA FT 50\%    & 0.0064 & 0.0152 & 0.0309 & 0.0427 & 0.0283 & 0.0320 & 0.0680 \\
				\midrule
				Mistral-7B & ICL Zeroshot    & 0.0090 & 0.7146 & 0.5391 & 0.6018 & 0.3823 & 0.5931 & 0.0005 \\
				& ICL Fewshot 1r  & 0.0090 & 0.0623 & 0.2080 & 0.2992 & 0.2160 & 0.3017 & 0.0001 \\
				& ICL Fewshot 5r  & 0.0090 & 0.0692 & 0.1691 & 0.1691 & 0.1628 & 0.1710 & 0.1986 \\
				& ICL Fewshot 10r & 0.0090 & 0.0563 & 0.1281 & 0.1450 & 0.1564 & 0.1708 & 0.2777 \\
				& LoRA FT 10\%    & 0.0084 & 0.0700 & 0.2170 & 0.2104 & 0.1163 & 0.1493 & 0.3884 \\
				\bottomrule
			\end{tabular}%
		\end{adjustbox}
	\end{table*}
	
	Transaction amount (\texttt{amt}) is well-reproduced across both models and all strategies, with TVD below 0.1 in nearly all configurations. The right-skewed shape of consumer spending amounts is recognizable enough that models can approximate it from schema context alone, without requiring example records.
	
	Geographic features (\texttt{lat}, \texttt{long}, \texttt{merch\_lat}, \texttt{merch\_long}) are where models consistently struggle to maintain numerical accuracy via ICL. Even at 10 examples, Qwen2.5-7B's longitude TVD remains at 0.2937; Mistral-7B reaches 0.1450 at the same example count. The problem is partly structural: US latitude--longitude pairs are bounded by land borders, and a handful of examples is nowhere near enough to convey that to the model. Only LoRA fine-tuning on Qwen2.5-7B brings geographic features within the 0.10 threshold.
	
	The \texttt{unix\_time} feature reveals an interesting asymmetry between the two models. Mistral-7B records TVD of 0.0005 at zero-shot ICL and 0.0001 at one-shot, while Qwen2.5-7B records 0.6931 at zero-shot ICL. However, the near-zero TVD for Mistral-7B most likely reflects the model defaulting to timestamps within a narrow range that happens to coincide with the real distribution's density peak, rather than any genuine modeling of the temporal distribution. If the model were genuinely learning the temporal distribution from examples, TVD should keep decreasing as more are added. Instead, it jumps from 0.0001 at one shot to 0.1986 at five, which suggests the additional examples pull the model away from a default that happened to work rather than actually improving its grasp of the temporal structure.
	
	Qwen2.5-7B LoRA fine-tuning achieves TVD below 0.10 for all seven numerical features at both data exposure levels, with the 50\% configuration showing modest but consistent improvement over 10\%.
	
	\paragraph{Categorical Distributional Fidelity}
	
	\begin{table}[tb]
		\centering
		\caption{Per-Feature Categorical TVD}
		\label{tab:categorical_tvd}
		\footnotesize
		\begin{adjustbox}{max width=\columnwidth}
			\begin{tabular}{llrrrr}
				\toprule
				\textbf{Model} & \textbf{Strategy} & \textbf{category} & \textbf{gender} & \textbf{job} & \textbf{state} \\
				\midrule
				Qwen2.5-7B & ICL Zeroshot    & 0.6694 & 0.1250 & 1.0000 & 0.7268 \\
				& ICL Fewshot 1r  & 0.6962 & 0.4308 & 0.9927 & 0.4516 \\
				& ICL Fewshot 5r  & 0.5324 & 0.4526 & 0.9840 & 0.3855 \\
				& ICL Fewshot 10r & 0.6396 & 0.4406 & 0.9837 & 0.3794 \\
				& LoRA FT 10\%    & 0.0486 & 0.0012 & 0.1551 & 0.0535 \\
				& LoRA FT 50\%    & 0.0474 & 0.0059 & 0.1430 & 0.0477 \\
				\midrule
				Mistral-7B & ICL Zeroshot    & 0.5528 & 0.0016 & 0.9999 & 0.7352 \\
				& ICL Fewshot 1r  & 0.5394 & 0.3488 & 0.9806 & 0.3515 \\
				& ICL Fewshot 5r  & 0.4293 & 0.2760 & 0.9593 & 0.3128 \\
				& ICL Fewshot 10r & 0.4870 & 0.3048 & 0.9542 & 0.3015 \\
				& LoRA FT 10\%    & 0.0687 & 0.1204 & 0.5943 & 0.2504 \\
				\bottomrule
			\end{tabular}
		\end{adjustbox}
	\end{table}
	
	The \texttt{job} field produces TVD near 1.0 under every ICL configuration for both models. In practice, this means the generated occupation labels have essentially no distributional overlap with the real data. Both models draw from a small vocabulary of common occupations that are well-represented in pre-training text but poorly representative of the actual dataset's distribution, which spans several hundred distinct occupations with a frequency structure that cannot be approximated from ten or fewer examples. Without weight updates~\cite{brown2020languagemodelsfewshotlearners}, ICL can bias generation toward in-context values but cannot reshape the prior over the full label vocabulary.
	
	LoRA fine-tuning on Qwen2.5-7B reduces \texttt{job} TVD to 0.1551 at 10\% exposure and 0.1430 at 50\%, still above the 0.10 threshold, but a substantial improvement over any ICL result. For \texttt{category}, \texttt{gender}, and \texttt{state}, both LoRA configurations achieve TVD below 0.06. Mistral-7B LoRA 10\% partially recovers \texttt{category} and \texttt{gender} but cannot close the gap on \texttt{job} (0.5943) or \texttt{state} (0.2504), suggesting that the 10\% training sample is insufficient to update the model's representation of high-cardinality and geographically structured features.

    %%% OLD %%%%%%%%%%%%%%%%%%%%%%%%%%%%%%%%%%%%%%%%%%%%%%%%%%%%%%%%%%%%%%%%%%
    % To determine where the observed failures on the \texttt{job} feature arise from dataset-specific artifacts or reflect a more general limitation of ICL, we construct a controlled synthetic benchmark isolating categorical distribution adaptation from other factors such as feature correlations, numerical structure, and domain semantics.
    %%%%%%%%%%%%%%%%%%%%%%%%%%%%%%%%%%%%%%%%%%%%%%%%%%%%%%%%%%%%%%%%%%%%%%%%%%

\begin{figure}[tb]
        \centering
        %\begin{subfigure}{0.45\columnwidth}
        \subfloat[ Mistral-7B-Instruct.\label{fig:tvd_mistral}]{%
        \includegraphics[width=.45\columnwidth]{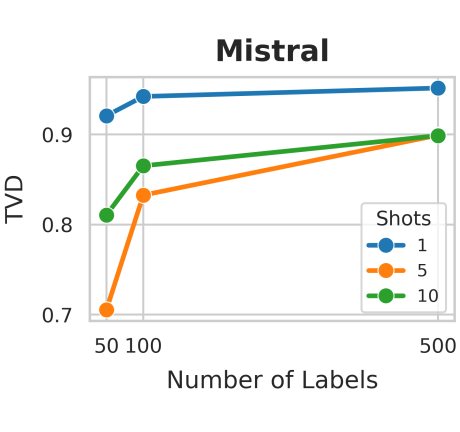}
        }\hfill
        \subfloat[Qwen2.5-7B-Instruct.\label{fig:tvd_qwen}]{\includegraphics[width=.45\columnwidth]{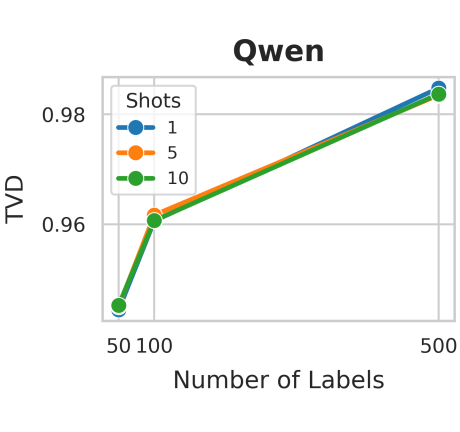}}
        \caption{TVD vs.\ cardinality. Zipf $\alpha=0.5$}
        \label{fig:tvd_cardinality}
    \end{figure}

    %%% NEW %%%%%%%%%%%%%%%%%%%%%%%%%%%%%%%%%%%%%%%%%%%%%%%%%%%%%%%%%%%%%%%%%%
    To determine whether the failures on \texttt{job} reflect a general ICL limitation, we evaluate TVD on synthetic Zipf-distributed features ($\alpha = 0.5$) across cardinalities of 50, 100, and 500, averaged over 30 runs.
    Most exemplary outcomes from the experimental results are presented in Figure~\ref{fig:tvd_cardinality}. % and~\ref{fig:tvd_qwen}, shot count has no measurable effect on categorical TVD for either model.
    One might notice in particular how in Figure~\ref{fig:tvd_qwen} all curves are virtually identical.

    %%%%%%%%%%%%%%%%%%%%%%%%%%%%%%%%%%%%%%%%%%%%%%%%%%%%%%%%%%%%%%%%%%%%%%%%%%
	
	\paragraph{Aggregate Fidelity}
	
	\begin{table}[tb]
		\centering
		\caption{Aggregate Mean TVD}
		\label{tab:aggregate_tvd}
		\footnotesize
		% \begin{adjustbox}{max width=\columnwidth}
			\begin{tabular}{llrrr}
				\toprule
				\textbf{Model} & \textbf{Strategy} & \textbf{Numerical TVD} & \textbf{Categorical TVD} & \textbf{Total TVD} \\
				\midrule
				Qwen2.5-7B & ICL Zeroshot    & 0.4892 & 0.6303 & 0.5405 \\
				& ICL Fewshot 1r  & 0.2972 & 0.6428 & 0.4229 \\
				& ICL Fewshot 5r  & 0.2268 & 0.5886 & 0.3584 \\
				& ICL Fewshot 10r & 0.2151 & 0.6108 & 0.3590 \\
				& LoRA FT 10\%    & 0.0383 & 0.0646 & 0.0479 \\
				& LoRA FT 50\%    & 0.0319 & 0.0610 & 0.0425 \\
				\midrule
				Mistral-7B & ICL Zeroshot    & 0.4058 & 0.5724 & 0.4664 \\
				& ICL Fewshot 1r  & 0.1566 & 0.5551 & 0.3015 \\
				& ICL Fewshot 5r  & 0.1355 & 0.4944 & 0.2660 \\
				& ICL Fewshot 10r & 0.1348 & 0.5119 & 0.2719 \\
				& LoRA FT 10\%    & 0.1657 & 0.2585 & 0.1994 \\
				\bottomrule
			\end{tabular}
		% \end{adjustbox}
	\end{table}
	
	For Qwen2.5-7B, numerical TVD decreases consistently across the few-shot ICL series, from 0.489 at zero-shot to 0.215 at 10 examples. Categorical TVD shows no comparable trend---it oscillates between 0.589 and 0.643 with no directional improvement as example count increases. Adding more examples improves the numerical features but has essentially no effect on the categorical ones. Switching to LoRA fine-tuning is not simply the next step in the same trend---the drop is much sharper: mean numerical TVD goes from 0.215 to 0.038, and mean categorical TVD from 0.611 to 0.065.
	
	For Mistral-7B, numerical TVD improves steadily through the few-shot series, reaching 0.135 at 10 examples---slightly better than Qwen2.5-7B at the same point. Fine-tuning, however, makes things worse: LoRA 10\% yields a numerical TVD of 0.166, which is actually higher than the best ICL result. On the categorical side, LoRA 10\% improves substantially over ICL (0.259 vs. 0.494--0.572), but remains far from the values Qwen2.5-7B achieves under fine-tuning.
	
	\subsection{Privacy and Memorization Risk}
	
	\begin{table}[tb]
		\centering
		\caption{Privacy Metrics}
		\label{tab:privacy}
		\footnotesize
		% \begin{adjustbox}{max width=\columnwidth}
			\begin{tabular}{llrr}
				\toprule
				\textbf{Model} & \textbf{Strategy} & \textbf{DCR Ratio} & \textbf{Exact Copies (\%)} \\
				\midrule
				Qwen2.5-7B & ICL Zeroshot    &  9.3597 & 0.00\% \\
				& ICL Fewshot 1r  &  4.4597 & 0.00\% \\
				& ICL Fewshot 5r  &  4.1291 & 0.00\% \\
				& ICL Fewshot 10r & 20.4266 & 0.10\% \\
				& LoRA FT 10\%    &  0.8655 & 0.10\% \\
				& LoRA FT 50\%    &  0.8609 & 0.05\% \\
				\midrule
				Mistral-7B & ICL Zeroshot    & 37.2999 & 0.00\% \\
				& ICL Fewshot 1r  &  3.6881 & 0.00\% \\
				& ICL Fewshot 5r  &  1.1806 & 0.00\% \\
				& ICL Fewshot 10r &  2.1070 & 0.00\% \\
				& LoRA FT 10\%    &  0.9512 & 0.00\% \\
				\bottomrule
			\end{tabular}
		% \end{adjustbox}
	\end{table}
	
	All ICL configurations yield DCR Ratios well above 1.0, ensuring synthetic records remain distant from the training data. The sole exception is Mistral-7B Fewshot 5r (1.1806), which approaches the memorization boundary. Conversely, all LoRA configurations fall below the 1.0 threshold; for Qwen2.5-7B, the proximity remains nearly identical at 10\% (0.8655) and 50\% (0.8609) exposure. This suggests that memorization signals saturate early rather than scaling with training volume. Notably, the exact copy rate for Qwen2.5-7B actually halves (from 0.10\% to 0.05\%) as exposure increases, confirming that distributional proximity and literal duplication are distinct phenomena. Mistral-7B LoRA 10\% similarly dips below the threshold (0.9512) without producing exact copies.
	
	\subsection{Fraud Class Reproduction} \label{sec:fraud_class_reproduction}
	
	Neither model produces a single fraudulent transaction under any ICL configuration---a clean failure on minority-class reproduction. LoRA fine-tuning partially restores the fraud class: Qwen2.5-7B produces 12 fraudulent records at 10\% exposure and 14 at 50\% (0.12\% and 0.14\% respectively), while Mistral-7B LoRA 10\% produces only 5 (0.05\%). All three fall well short of the real distribution's 0.58\% base rate, meaning the class imbalance is substantially underrepresented even after fine-tuning.
	
	\subsection{Inter-Feature Correlation Preservation}
	
	Since TVD captures only marginal distributions, we assess dependency structures by computing the correlation matrix across the seven numerical features and the binary fraud label. We report two summary statistics in Table \ref{tab:correlation}: the maximum absolute pairwise correlation error (the most discrepant pair) and the total absolute correlation error (the sum of all off-diagonal absolute errors in the $8 \times 8$ matrix). These metrics, evaluated for Qwen2.5-7B across zero-shot, few-shot (10r), and LoRA (50\%) configurations, reveal whether synthetic datasets preserve the real data's underlying feature interactions.
	
	\begin{table}[tb]
		\centering
		\caption{Inter-Feature Correlation Errors (Qwen2.5-7B)}
		\label{tab:correlation}
		\footnotesize
		% \begin{adjustbox}{max width=\columnwidth}
			\begin{tabular}{lrrl}
				\toprule
				\textbf{Strategy} & \textbf{Max Error} & \textbf{Total Error} & \textbf{Worst Pair} \\
				\midrule
				ICL Zeroshot    & 0.40 & 4.34 & lat/long, merch\_lat/merch\_long \\
				ICL Fewshot 10r & 0.12 & 2.26 & lat/long, merch\_lat/merch\_long \\
				LoRA FT 50\%    & 0.15 & 0.82 & amt / is\_fraud \\
				\bottomrule
			\end{tabular}
		% \end{adjustbox}
	\end{table}
	
\begin{figure}[tb]
		\centering
		\begin{subfigure}{0.45\textwidth}
			\includegraphics[width=\linewidth]{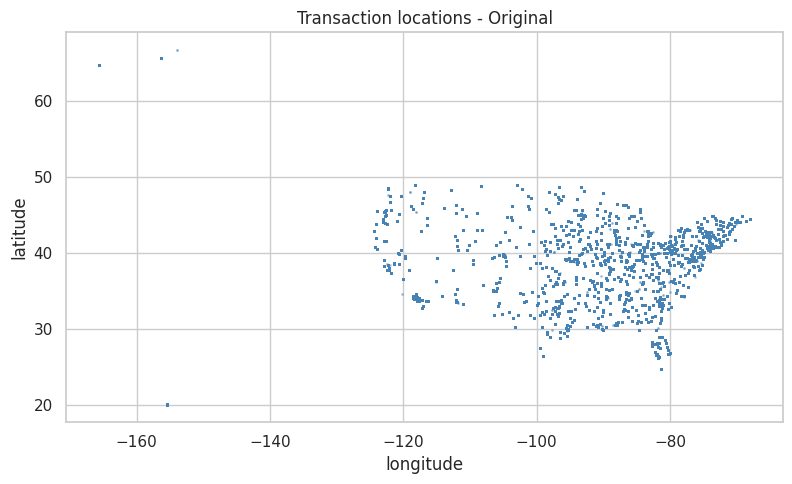}
			\caption{Original}
		\end{subfigure}
		\hfill
		\begin{subfigure}{0.45\textwidth}
			\includegraphics[width=\linewidth]{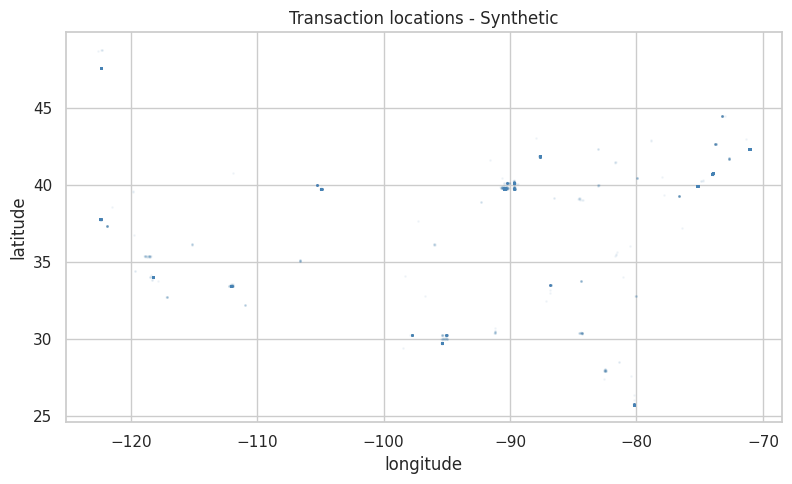}
			\caption{Zero-shot}
		\end{subfigure}
		\\
		\begin{subfigure}{0.45\textwidth}
			\includegraphics[width=\linewidth]{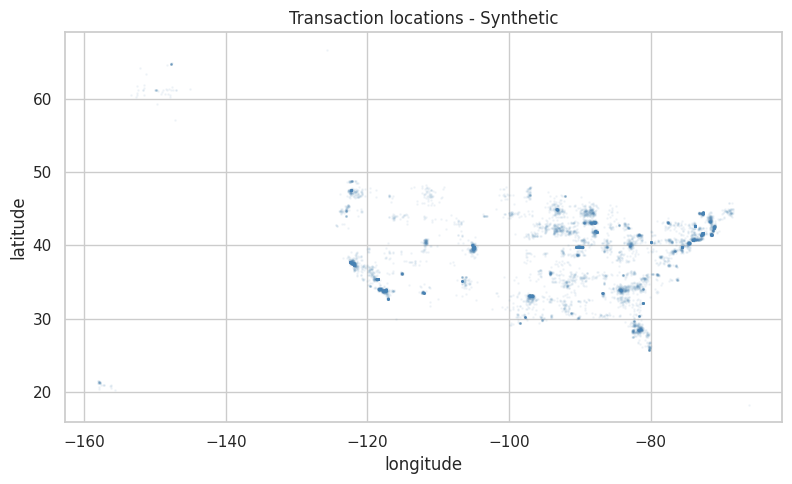}
			\caption{Few-shot 10r}
		\end{subfigure}
		\hfill
		\begin{subfigure}{0.45\textwidth}
			\includegraphics[width=\linewidth]{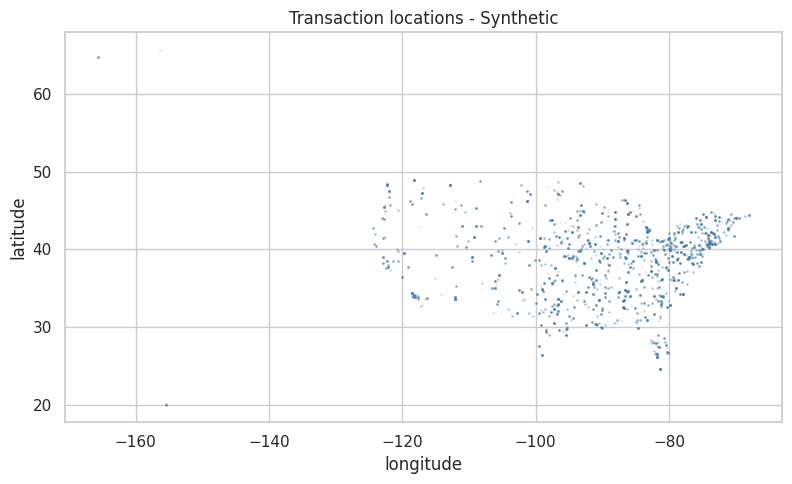}
			\caption{LoRA 50\%}
		\end{subfigure}
        \caption{Geographic distribution across generation strategies (Qwen2.5-7B)}
        \label{fig:geodistribution}
    \end{figure}

\begin{figure}[tb]
		\begin{subfigure}{0.45\textwidth}
			\includegraphics[width=\linewidth]{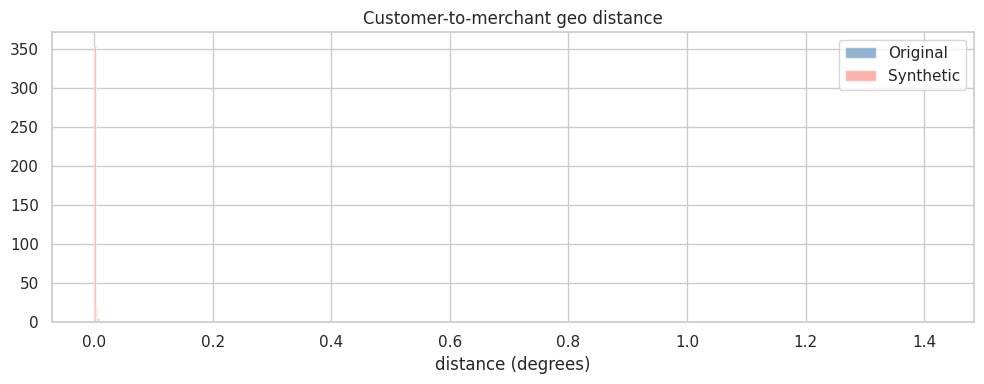}
			\caption{Zero-shot}
		\end{subfigure}
		\hfill
		\begin{subfigure}{0.45\textwidth}
			\includegraphics[width=\linewidth]{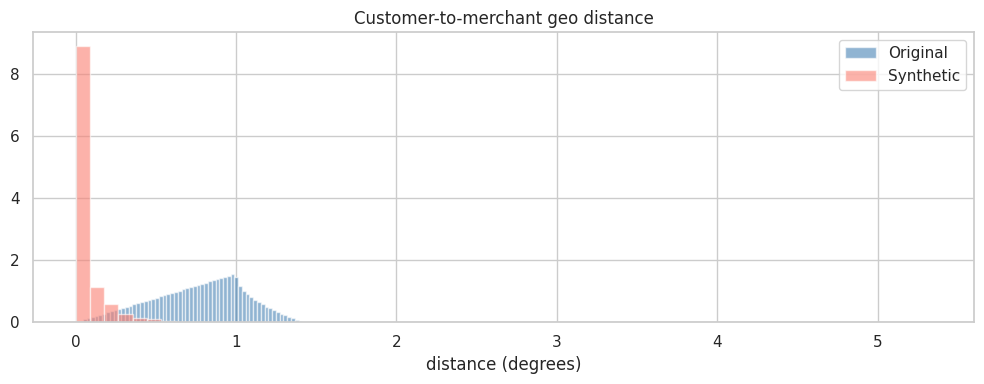}
			\caption{Few-shot 10r}
		\end{subfigure}
	    \\
		\begin{subfigure}{0.45\textwidth}
			\includegraphics[width=\linewidth]{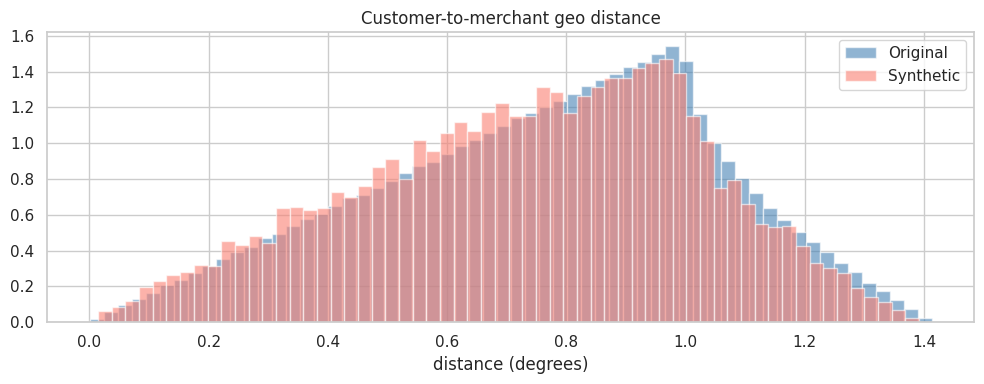}
			\caption{LoRA 50\%}
		\end{subfigure}

		\caption{Geographic distance across generation strategies (Qwen2.5-7B)}
		\label{fig:geodistance}
	\end{figure}

\begin{figure}[tb]
        \begin{subfigure}{0.45\textwidth}
			\includegraphics[width=\linewidth]{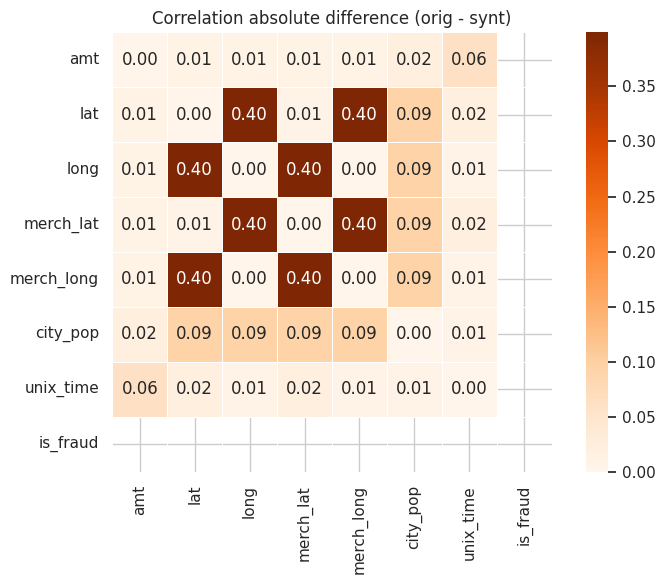}
			\caption{Zero-shot}
		\end{subfigure}
		\hfill
		\begin{subfigure}{0.45\textwidth}
			\includegraphics[width=\linewidth]{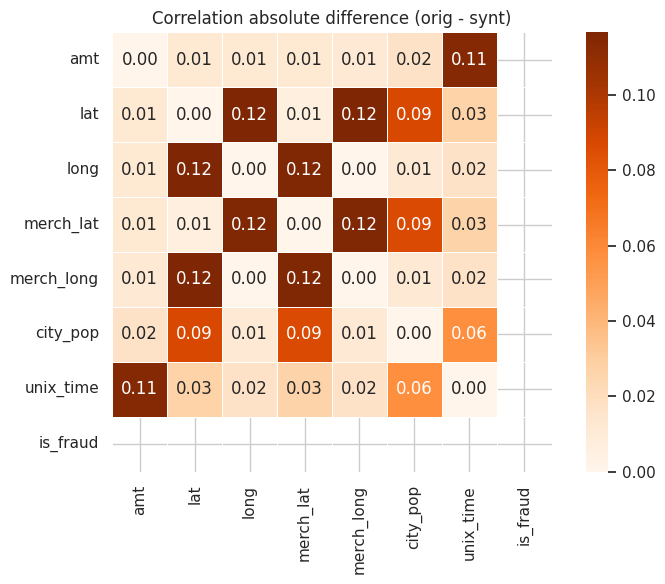}
			\caption{Few-shot 10r}
		\end{subfigure}
		\\
		\begin{subfigure}{0.45\textwidth}
			\includegraphics[width=\linewidth]{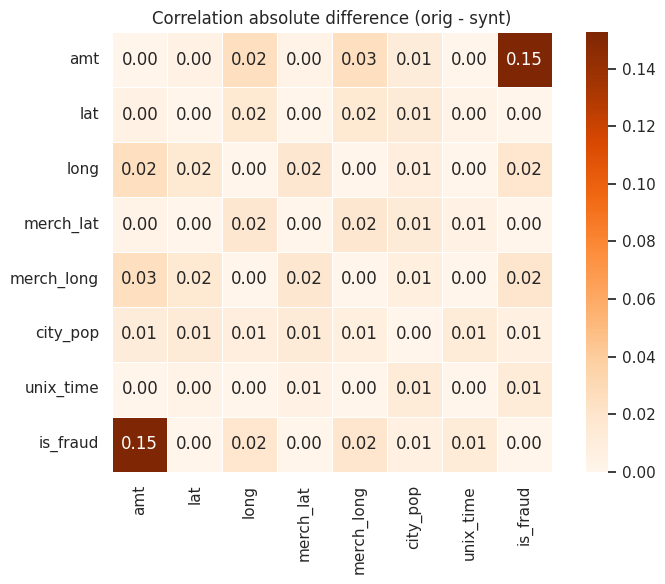}
			\caption{LoRA 50\%}
		\end{subfigure}
		\caption{Correlation heatmap across generation strategies (Qwen2.5-7B)}
		\label{fig:heatmap}
	\end{figure}

	Under zero-shot ICL, the most severe failures concentrate in the geographic block. The correlation between cardholder and merchant coordinates ($r > 0.99$ in real data) collapses to near-zero in synthetic output, reaching a maximum pairwise error of 0.40. As shown in Fig. \ref{fig:geodistribution}, zero-shot models generate coordinates independently, lacking the spatial co-occurrence typical of legitimate transactions. Providing ten in-context examples halves the total correlation error (from 4.34 to 2.26) and reduces the geographic error to 0.12, suggesting partial inference of spatial coupling. However, Fig. \ref{fig:geodistance} and \ref{fig:geodistribution} reveal that few-shot generation still fails to replicate co-location structure, instead clustering around urban centers. These issues, consistent with the high sensitivity of geographic TVD noted in Sec. \ref{sec:numerical_distributional_fidelity}, are resolved only via LoRA fine-tuning.
    LoRA FT (50\% exposure) reduces total correlation error to 0.82---five times lower than zero-shot---and eliminates geographic discrepancies (Fig. \ref{fig:heatmap}). The worst-case error shifts to the \texttt{amt}/\texttt{is\_fraud} pair (0.15); while the model learns to produce fraudulent records, it lacks the full calibration between high amounts and fraud present in real data. This aligns with the fraud underrepresentation in Sec. \ref{sec:fraud_class_reproduction}, as a deficit in minority-class samples inherently weakens the captured correlation. Consequently, correlation fidelity serves as a vital diagnostic for structural dependencies that marginal TVD statistics overlook.
	
%% file: discussion.tex
\section{Discussion}
\label{sec:discussion}

\paragraph{Categorical Prior Lock-in in In-Context Learning}
% \label{sec:prior-lock-in}

The central finding of this work is that in-context learning (ICL) exhibits a structural limitation when applied to structured generation tasks requiring adaptation to new categorical distributions. Across all ICL configurations and both models, high-cardinality categorical features remain poorly approximated, with total variation distance (TVD) approaching 1.0 regardless of the number of in-context examples. Increasing the number of examples improves numerical fidelity but does not produce any consistent improvement in categorical distributions, and minority categories are never reproduced.

As already pointed out in Section \ref{sec:intro}, we refer to the tendency of a language model to adhere to its pre-training distribution over tokens (even when conditioned on examples drawn from a different target distribution) as \textit{categorical prior lock-in}. In the case of the \texttt{job} feature, both models repeatedly generate a small set of common occupations that are likely overrepresented in pre-training corpora, while failing to approximate the long-tailed empirical distribution present in the dataset.

This behavior reflects a fundamental property of ICL. While ICL conditions generation on a finite set of examples, it does not update the model’s parameters and therefore cannot globally reshape the underlying token probability distribution. As a result, ICL can bias generation toward values observed in the prompt, but cannot reweight the full support of a high-cardinality categorical variable. When the target distribution differs significantly from the model’s pre-training prior---as is typical for domain-specific structured data---this leads to systematic mismatch.

This limitation is not resolved by prompt design or example scaling within the tested range. Instead, it appears to be inherent to the mechanism of conditioning without learning: ICL operates locally in context, while the task of approximating a high-cardinality distribution requires global adjustment of token probabilities. Our results suggest that this gap cannot be closed without parameter updates.

\paragraph{Beyond Marginals: Structural Failures of ICL}

The effects of categorical prior lock-in extend beyond marginal distributions. Under zero-shot ICL, geographic features are generated independently, collapsing strong correlations present in the real data; few-shot prompting partially reduces these errors but does not recover the underlying dependency structure. Because categorical variables often serve as anchors for inter-feature dependencies, failure to model their distribution propagates to the broader joint structure of the data. These results suggest that ICL is insufficient for structured generation tasks requiring both accurate marginals and coherent joint distributions under distribution shift.

\paragraph{Fine-Tuning as Distributional Adaptation}

In contrast to ICL, parameter-efficient fine-tuning (LoRA) substantially improves both categorical and numerical fidelity. For Qwen2.5-7B, fine-tuning reduces categorical TVD by an order of magnitude and restores inter-feature correlations that are absent under ICL. These results support the interpretation that weight updates enable the model to adapt its internal representation of token distributions to match the target dataset.

From this perspective, fine-tuning can be understood as performing the global distributional adjustment that ICL cannot. By modifying model parameters, fine-tuning reshapes the probability mass assigned to tokens across the full vocabulary, allowing the model to represent high-cardinality categorical distributions more accurately.

However, this improvement comes at a cost. All fine-tuned configurations exhibit DCR ratios below 1.0, indicating that synthetic records are, on average, closer to real records than real records are to each other. This suggests the emergence of memorization behavior, even at relatively low data exposure levels. The trade-off is therefore clear: while fine-tuning enables distributional adaptation, it introduces risks associated with data leakage and reduced privacy.

\paragraph{Instability of Fine-Tuning in Small Instruction-Tuned Models}
% \label{sec:instability}
 
A key finding of our experiments is that fine-tuning can degrade instruction-following in small models. The behaviour of Mistral-7B under LoRA fine-tuning illustrates this risk concretely: unlike Qwen2.5-7B, which maintains stable structured output generation after adaptation, Mistral-7B exhibits severe degradation in output validity, culminating in complete generation failure at higher data exposure levels---producing no valid records whatsoever in 1\,000 attempts at 50\% data exposure. This is not a marginal degradation but a catastrophic collapse of the model's ability to follow the structured output schema it had previously adhered to under ICL. This instability aligns with prior evidence that fine-tuning small models can be harmful, potentially because specialized adaptation saturates limited model capacity \cite{wei2022finetunedlanguagemodelszeroshot} or induces catastrophic forgetting of core instruction-following abilities during domain specialization \cite{fatemi2024comparativeanalysisinstructionfinetuning}.

One possible interpretation is that the additive updates introduced by LoRA interfere with the latent representations responsible for schema adherence. While the base weights remain frozen, the injected low-rank signal may disrupt the pre-existing instruction-following logic. Small models with less representational redundancy, such as Mistral-7B, appear particularly vulnerable to this low-rank interference, which effectively overrides the model's structural priors. These results indicate that output stability depends on the compatibility between the adapter's signal and the base model's architecture, a factor that must be evaluated alongside standard performance metrics.

\paragraph{Implications for LLM-Based Structured Generation}

The findings of this work point to a fundamental trade-off in the use of LLMs for structured data generation. In-context learning offers a lightweight and privacy-preserving mechanism for adapting models to new tasks, but is fundamentally limited in its ability to approximate distributions that diverge from pre-training priors. Fine-tuning overcomes these limitations by enabling global adaptation, but introduces risks related to memorization and, in some cases, model instability.

More broadly, these results suggest that ICL should not be interpreted as a general-purpose substitute for learning in settings that require distributional adaptation. Instead, its strengths lie in tasks that align closely with pre-training distributions or require only local adjustments. For structured generation tasks involving high-cardinality categorical variables or strong distribution shift, parameter updates appear to be necessary.

This perspective reframes the role of ICL within the broader landscape of LLM adaptation methods: rather than a universal mechanism for few-shot learning, it is a form of conditional generation with inherent limitations determined by the model’s prior. Understanding these limitations is essential for deploying LLMs in real-world settings where distribution mismatch is the norm rather than the exception.

%% file: conclusion.tex
\section{Conclusion}
We investigated the viability of 7B-scale language models for synthetic tabular data generation, comparing in-context learning (ICL) and LoRA fine-tuning across two models (Qwen2.5-7B and Mistral-7B) on a fraud detection dataset. Our results show that the generation paradigm—rather than model choice—is the primary determinant of data quality at this scale. ICL proves insufficient for high-cardinality categorical features: marginal fidelity on \textit{job} remains near-random regardless of shot count or model, and neither model generates fraudulent transactions under any ICL configuration, rendering ICL-produced data unsuitable for fraud detection tasks. Numerical fidelity and inter-feature correlation structure improve with more in-context examples but plateau well below the quality achievable through fine-tuning. LoRA fine-tuning on Qwen2.5-7B substantially closes these gaps, recovering both marginal categorical distributions and geographic correlation structure. However, DCR analysis reveals a memorization signal that persists across data exposure levels, raising concerns about the privacy guarantees of the generated data. Mistral-7B proves unstable under LoRA adaptation at the configurations tested, failing to produce valid outputs at higher data exposure. 

Future work will focus on validating these insights across a broader range of tabular domains and incorporating formal downstream utility evaluations. Additionally, we aim to integrate differentially private fine-tuning mechanisms (e.g., DP-LoRA) into the pipeline to formally mitigate the identified memorization risks while preserving structural and distributional fidelity.